\documentclass{article}
\usepackage{spconf,amsmath,graphicx}
\usepackage{hyperref}
\usepackage{xcolor}
\usepackage{multirow}

\title{Self-supervised Learning with Local Attention-Aware Feature}

\name{ Trung Pham  $^{\star}$ \qquad Rusty John Lloyd Mina $^{\star}$ \qquad Dias Issa $^{\star}$ \qquad Chang D. Yoo \thanks{$^*$ These authors have equal contributions. The order of the authors has been randomly selected by applying the rock, scissors, paper based challenge.}} 
\address{Korea Advanced Institute of Science and Technology (KAIST)}

\newcommand{\etal}{\textit{et al. }}

\begin{document}
\maketitle
\begin{abstract}
In this work, we propose a novel methodology for self-supervised learning for generating global and local attention-aware visual features. Our approach is based on training a model to differentiate between specific image transformations of an input sample and the patched images. Utilizing this approach, the proposed method is able to outperform the previous best competitor by 1.03\% on the Tiny-ImageNet dataset and by 2.32\% on the STL-10 dataset. Furthermore, our approach outperforms the fully-supervised learning method on the STL-10 dataset. Experimental results and visualizations show the capability of successfully learning global and local attention-aware visual representations. 
\end{abstract}


\section{Introduction}
\label{sec:intro}

Visual representation learning is one of the most important research directions in the fields of computer vision and image processing. Pre-training deep neural networks on large labeled datasets, such as ImageNet~\cite{deng2009imagenet}, with the purpose of learning useful generic features, enhances their performances on downstream tasks. However, the main limitation of this method stems from the necessity for a large manually labeled dataset, which in turn requires human labor, and thus, it is costly and prone to errors.

Nevertheless, in recent years, various solutions were proposed to overcome this limitation. One promising approach is self-supervised learning (SSL) which enables learning image features using only visual information without dependency on manually provided annotations. The main idea of this approach is to get surrogate supervision by solving {\it pretext} tasks on the input data used during the pre-training stage. While solving the pretext tasks, neural network learns useful visual representations that can be used for other downstream computer vision problems ~\cite{bachman2019learning}. Different pretext tasks were proposed in previous works, including prediction of the rotation angle~\cite{gidaris2018unsupervised}, prediction of the relative positioning of image patches~\cite{doersch2015unsupervised}, or image colorization~\cite{zhang2016colorful}. 
Despite the strong performance of supervised learning methods using large manually labeled datasets, self-supervised learning methodologies are confidently closing the gap during recent years~\cite{gidaris2018unsupervised, bachman2019learning, caron2018deep}. Furthermore, recently proposed SSL methodologies achieve state-of-the-art results and outperform supervised learning methods in some of the tasks~\cite{caron2018deep, hjelm2018learning}.  

\begin{figure}[t]
    \centerline{\includegraphics[width=1\linewidth]{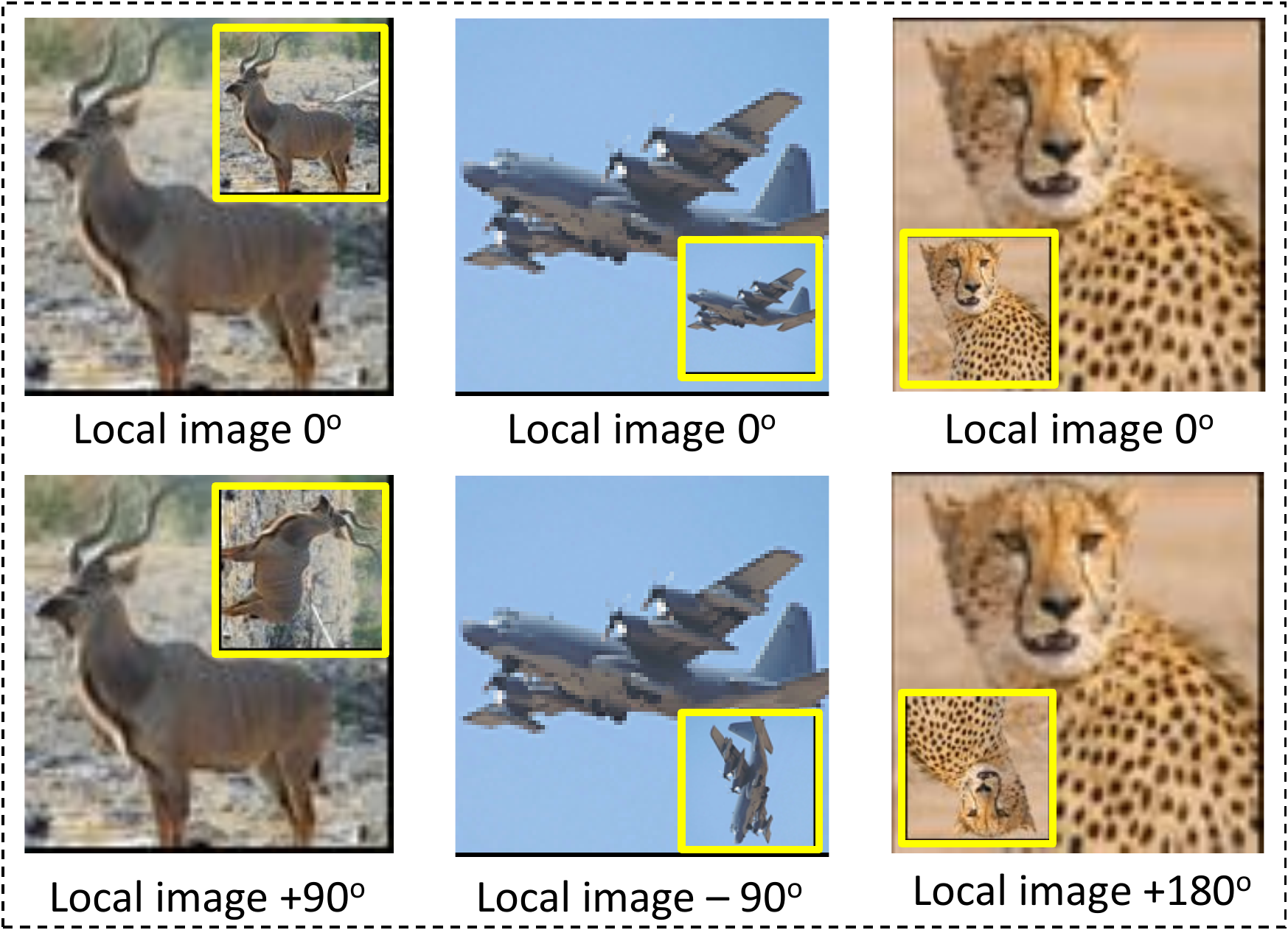}}
    \caption{Can the model understand both global context and local details to recognize how the patch changed (i.e. predict the patch rotation)? }
    \label{fig:res_0}
\end{figure}

\begin{figure*}[ht!]
\vspace{0.1cm}
\centerline{\includegraphics[width=1\linewidth]{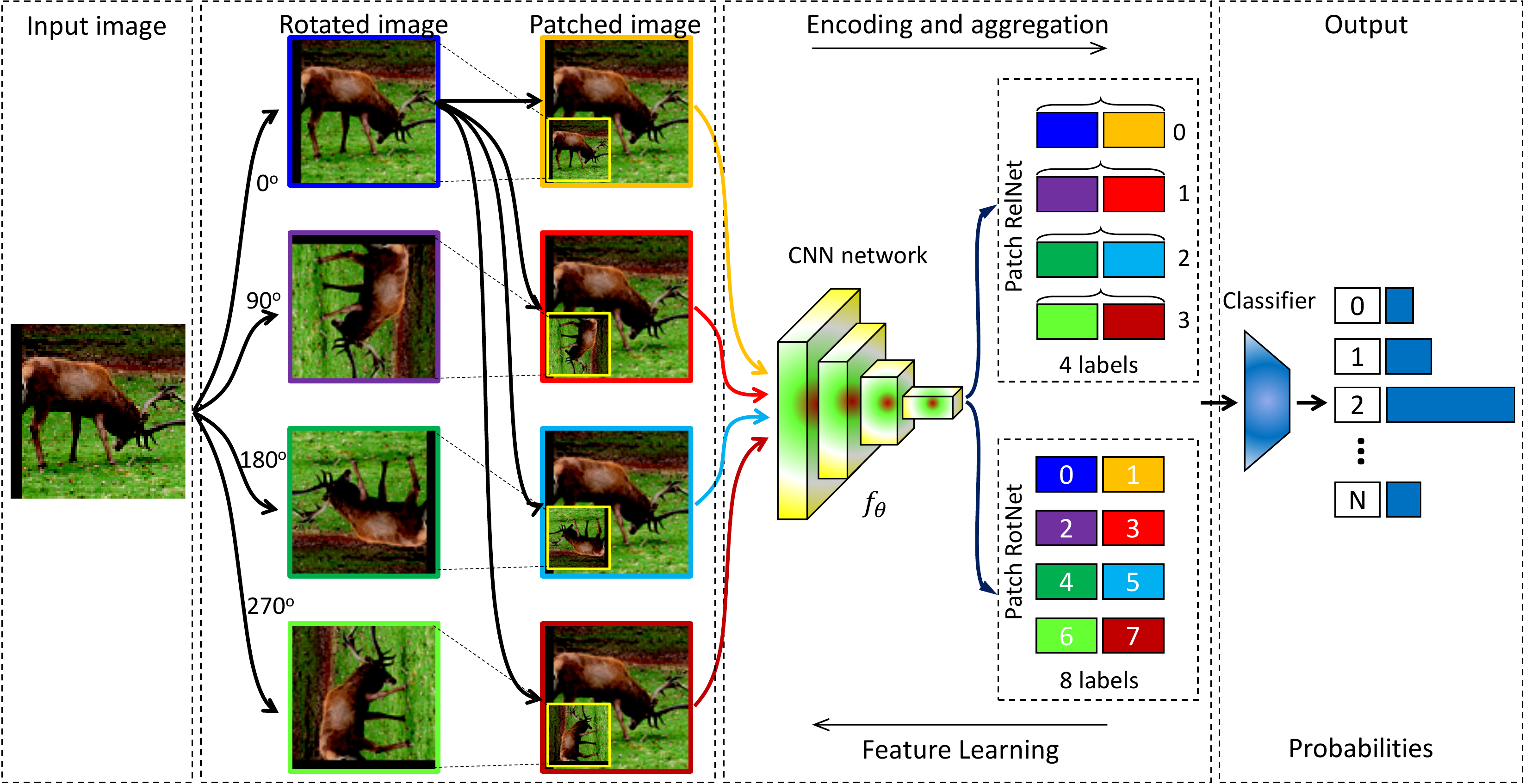}}
\caption{\textbf{Detailed architecture:} to achieve a good visual representation of the given image, the model learns to understand both global (the outer image) and local contexts (small patch with yellow border) so that it can predict the rotation angle of a patched image.}
\label{fig:res_1}
\end{figure*}

Inspired by the success of previously proposed self-supervised learning methods, in this paper, we propose a novel pretext task for self-supervised learning. Our pretext task (Fig.~\ref{fig:res_0}) consists of 3 steps: 1) producing a rotated image by rotating an unlabeled original input image by 0, 90, 180, or 270 degrees; 2) producing a patch by resizing the rotated image; 3) predicting the rotation angle of rotated image or rotated patch image. 
Our main contributions are summarized as follows:\vspace{-0.2cm}
\begin{itemize}
    \item We propose a novel pretext task for learning global and local-aware for better visual representations.\vspace{-0.2cm}
    \item We introduce two SSL approaches for solving above pretext task.\vspace{-0.2cm}
    \item We demonstrate the effectiveness of the proposed methodology in comparison with previous state-of-the-art methods on two standard benchmark datasets.
\end{itemize}


\section{Related works}
\label{sec:rel_works}
Self-supervised learning (SSL) was utilized for different domains including video~\cite{wang2020self, Tschannen_2020_CVPR, vondrick2018tracking}, audio ~\cite{alwassel2020self, zhao2018sound, gao2018learning}, and image~\cite{zhang2017split, larsson2017colorization, mundhenk2018improvements}. In this work, we focus on the latter.
There is a wide range of works for SSL proposed previously that are based on pretext tasks utilizing images. Doersch \etal~\cite{doersch2015unsupervised} introduced a pretext task that requires prediction of the relative position of patches extracted from an image. After pre-training the model with this pretext task, the authors evaluated the performance of their model by using the bounding box prediction task on Pascal Visual Object Classes (PASCAL-VOC) challenge dataset~\cite{everingham2010pascal}. Despite achieving significant enhancement in performance compared to a randomly initialized model, their results still do not reach the level of supervised pre-training on ImageNet~\cite{deng2009imagenet} labels. 

In a work by Noroozi and Favaro~\cite{noroozi2016unsupervised}, the pretext task is defined as solving the jigsaw puzzles of the input images by training the model to arrange randomly shuffled patches of a given image. Their results show that they were able to outperform previous works, including Doersch \etal ~\cite{doersch2015unsupervised}.
In contrast to the previous works, Zhang \etal~\cite{zhang2016colorful}, utilized  the color information of a given image instead of its spatial evidence extracted from image patches. They defined the pretext task as the colorization of a gray-scale input image. Utilizing this approach, Zhang \etal were able to outperform previous works on PASCAL-VOC~\cite{everingham2010pascal} and ImageNet~\cite{deng2009imagenet} datasets.

Gidaris \etal~\cite{gidaris2018unsupervised} defined their pretext task as the prediction of the rotation angle of an input image. The rationale behind this is the network should recognize the objects shown in the image to determine the image's rotation angle. With this approach the authors were able to achieve significant performance gain on several datasets including CIFAR-10~\cite{krizhevsky2009learning}, PASCAL-VOC~\cite{everingham2010pascal} and ImageNet~\cite{deng2009imagenet}.

In a work by Caron \etal~\cite{caron2018deep} proposed a pretext task that works with image features rather than with images themselves. The authors utilize k-means clustering algorithm on visual features extracted from input images. The cluster assignments served as pseudo-labels for self-supervised classification. These pseudo-labels are utilized in training a classifier on top of the feature extraction network. This approach allowed the authors to outperform previous works on multiple datasets, such as Places~\cite{zhou2014learning}, PASCAL-VOC~\cite{everingham2010pascal} and ImageNet~\cite{deng2009imagenet}.

Hjelm \etal~\cite{hjelm2018learning} introduced the Deep InfoMax approach as a pretext task for self-supervised learning. The proposed idea is to maximize the mutual information between global and local features extracted from an input image. The authors were able to perform better than a number of previous works while showing superior results in some tasks in comparison with supervised learning on such datasets as CIFAR-10, CIFAR-100~\cite{krizhevsky2009learning}, STL-10~\cite{coates2011analysis}, ImageNet~\cite{deng2009imagenet}. 

Here we propose to classify the patched images (which are auto-generated from a given image) to encourage the learner to look simultaneously into both the entire image (global context) and the small patch within the image to predict what is changed in the small patch (local content) as shown in Fig. \ref{fig:res_0} and Fig.~\ref{fig:res_3} (with attention map of GradCAM visualization).

\section{Method}
\label{sec:method}
We begin our method by designing a new pretext task based on the local and global context-aware features as depicted in Fig.~\ref{fig:res_1}. Given an image $X$, the transformation $T(X)$ is applied to the input image to generate four new images. Here $T$ is the \textit{rotate} operation (and in some case, \textit{resize-and-paste} operation). We follow Gidaris \etal ~\cite{gidaris2018unsupervised} method and use multiples of $90^o$ as angel of rotations, which give us four images that are rotated by multiple of $90^o$: $0^o,$ $90^o,$ $180^o,$ $270^o$. Each of the rotated image is resized to a smaller patch (using bilinear interpolation) and pasted as patches with original upright image ($0^o$) as the background; generating four patched images. 

\begin{figure}[t]
    \centerline{\includegraphics[width=1\linewidth]{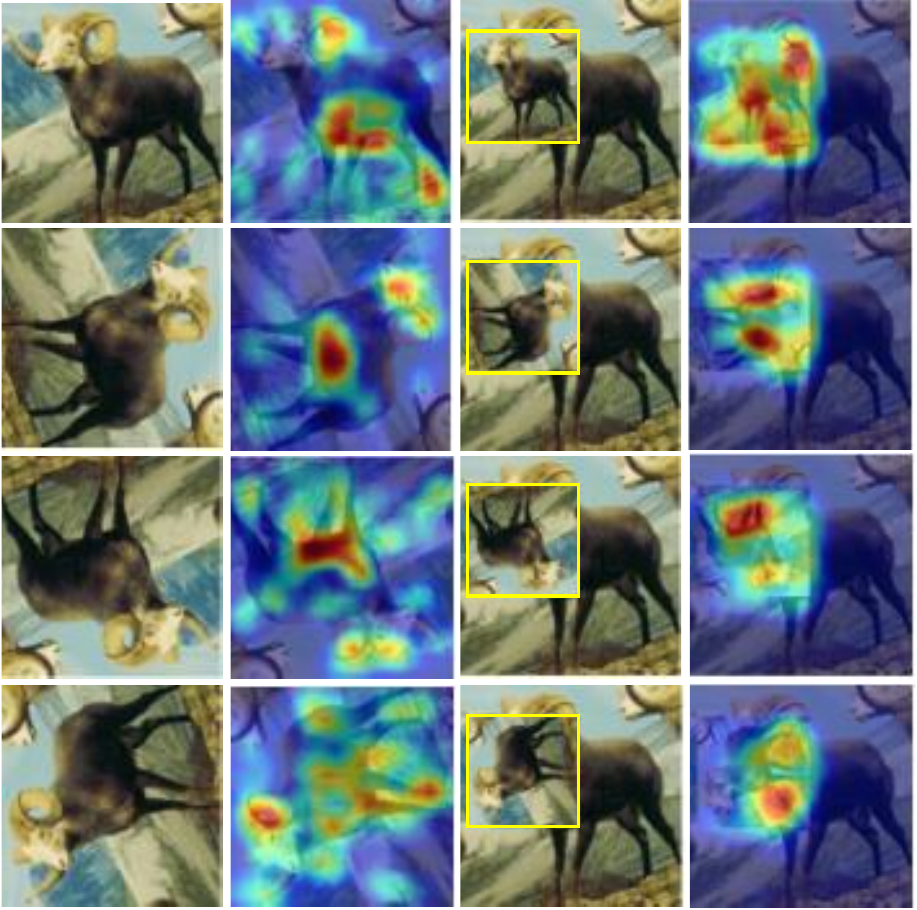}}
    \caption{GradCAM visualization of the our proposed method for the 8 generated images with the ``Patch RotNet'' learning.}
    \label{fig:res_3}
\end{figure}

The resize ratio is determined based on empirical experiments (we found 0.4 gives the best performance), while the position to paste the patch into is randomly selected for each input image (Fig.~\ref{fig:res_0}). This encourages the network to not only look into a specific region when predicting but also to differentiate the region whether it belongs to local features or global features (Fig. \ref{fig:res_3}). 
We argue that to classify the four patched images, the learner must be aware of the global context and the local context. Finally, we define the pretext task for self-supervision as predicting the angle of rotation of either the image and/or the rotation patch (Fig. \ref{fig:res_1}). By solving this pretext task, the learner learns good visual representations for each input image. 
The resulting images are encoded by a CNN network to produce a 64-dimension vector followed by a linear layer to classify into four or eight of possible classes. Given $L$ unlabeled images for training $(x_1, x_2, ...,x_L)$, the transformed images $\{T_j\}_{j=1,...,M}$ (here $M$ = 8). The learner $f_{\theta}$ is trained to predict the transformation applied to the image. In the case of rotated original image, we predict the angle of rotation of the original image, and in the case of patched image, we predict the angle of rotation of the rotated patch. We train the network $f_{\theta}$ by minimizing the following self-supervised learning objective function:
\begin{equation}
    \mathcal{L}_{SSL}(\{T_j\}_{j=1,...,M}) = \min_{\theta}\frac{1}{NL}\sum_{i=1}^{L}\sum_{y=0}^N \ell \left( f_{\theta} (T_y(x_i)), y \right),
\end{equation}
where $\ell$ denotes the cross-entropy loss, $N$ is number of class. After unsupervised training with the aforementioned task on the unlabeled dataset, we use the pretrained model to evaluate on the labeled dataset. Based on the proposed method, we introduce two SSL approaches to show the advantages of each of them. 
According to the first SSL, given an input image, 4 rotated images are generated ($0^o, 90^o, 180^o, 270^o$ as suggested in~\cite{gidaris2018unsupervised}). Keeping the original image, the rotated images are resized to a smaller image with a chosen ratio and pasted into the original image to form 4 patched images (Fig.~\ref{fig:res_1}). This approach contains 8 images as input into the CNN network for feature learning, which we refer to as ``Patch RotNet'' network (softmax output with N = 8 classes, labels from 0,...,7). 

The second proposed network contains the 8 above images, but rather than feeding each image as one sample with one label into the network, we aggregate images into pairs: the rotated image and the patched image (Fig.~\ref{fig:res_1}). Each image in each pair is encoded into the latent vector and then is aggregated by one relation module $\alpha = \text{concat}(f_{\theta}(x_1),f_{\theta}(x_2))$, here we use the concatenation operator to model the relation between the two images in the pairs. We refer this approach as ``Patch RelNet'' (softmax output with N = 4 classes).

\section{Experiments}
\label{sec:exps}
To verify the effectiveness of our proposed method and compare to other SSL schemes, we use two challenging datasets: \\
\textbf{1.~Tiny-ImageNet}~\cite{krizhevsky2009learning}: a smaller version of the ImageNet dataset with 200 classes, images is scaled to $64\times 64$ pixels. \\
\textbf{2.~STL-10}~\cite{coates2011analysis}: image recognition dataset with 10 classes designed for unsupervised learning purposes. It contains 500 labeled images for each class and 100k unlabeled images. Images were drawn from ImageNet and scaled to $96\times 96$.
\begin{figure}[t]
    \centerline{\includegraphics[width=1\linewidth]{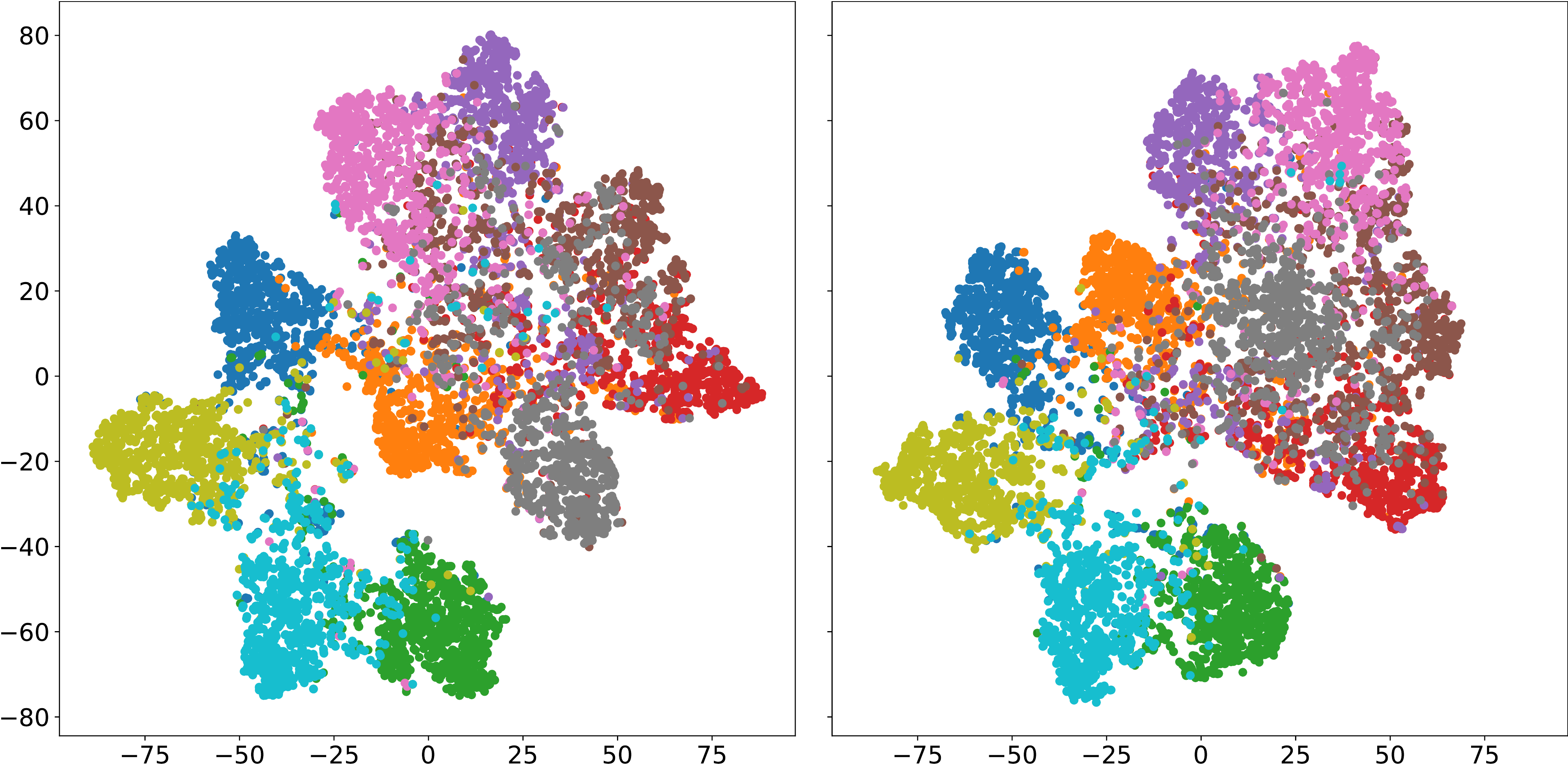}}
    \caption{t-SNE plot of the proposed method (left side) and RotNet (right side). The features were extracted from the top layer of the ResNet32 models finetuned on STL-10 dataset.}
    \label{fig:res_4}
\end{figure}
\begin{table}[ht!]
    \centering
    \begin{tabular}{|c|c|c|c|c|} \hline
    \multirow{2}{*}{Method} & \multicolumn{2}{c|}{Tiny} & \multicolumn{2}{c|}{STL-10} \\ \cline{2-5} 
     & Linear & Tuned & Linear & Tuned \\ \hline \hline
    Supervised & \multicolumn{2}{c|}{40.45} & \multicolumn{2}{c|}{73.8} \\ \hline
    Random weight & 6.63 & 36.78 & 26.66 & 57.96 \\ \hline
    Deep InfoMax~\cite{hjelm2018learning} & 14.90 & 27.78 & 41.64 & 57.66  \\ \hline
    Deep Cluster~\cite{caron2018deep} & 12.57 & 27.14 & 41.70 & 59.05 \\ \hline
    RotNet~\cite{gidaris2018unsupervised} & 16.32 & 29.07 & \textbf{60.48} & 68.16 \\ \hline \hline
    \textbf{Patch RelNet 0.3} & \textbf{17.35} & \textbf{29.32} & 59.09 &  \textbf{69.95}\\ 
    \textbf{Patch RelNet 0.4} & 16.04 & \textbf{29.19} & \textbf{60.28} & \textbf{70.08} \\ 
    \textbf{Patch RelNet 0.5} & 14.50  & 28.54 & 59.54 & \textbf{69.81}\\ 
    \textbf{Patch RotNet 0.3} & 13.15& \textbf{30.01} & 55.23  & \textbf{69.17}\\ 
    \textbf{Patch RotNet 0.4} & 11.70 & 28.90 & 54.55  & \textbf{69.28} \\ 
    \textbf{Patch RotNet 0.5} & 12.45 & \textbf{29.29} & 57.55 & \textbf{69.43} \\
    \hline
    \end{tabular}
    \caption{Test accuracy. Self-supervised training on unlabeled data with linear evaluation and finetuning on labeled data. Experiments are done on two challenging datasets: STL-10 and Tiny-ImageNet with a shallow backbone (\textcolor{blue}{ResNet-8}). \textbf{Best results are highlighted in bold}.}
    \label{tab:resnet8}\vspace{-0.2cm}
\end{table}
We evaluate the proposed self-supervised learning method by linear evaluation and finetune experiments. First, we trained self-supervised learning (SSL) for 200 epochs (without using any labels), batch size set as 64. In the linear evaluation, the model which is pretrained in SSL is applied as the feature extractor (all layers are frozen) to extract features for images in the test set, and train for 100 epochs with two fully-connected layers for classification, and evaluation in the test set with the labels. In the finetune experiments, the feature extractor pretrained in SSL manner is finetuned, together with the two fully-connected layer for classification is finetuned for 20 epochs. 
All experiments for linear evaluation and finetune use the batch size 32. The proposed method has been compared against other SSL methods in the two different CNN architecture including ResNet8 (shallower) and ResNet32 (deeper).

The ratio of the small patch is set as 0.3, 0.4, 0.5 for both two proposed networks. We compare our proposed method with various methods including supervised learning, network initialized with random weights, and three strong competitors such as Deep InfoMax~\cite{hjelm2018learning}, Deep Cluster~\cite{caron2018deep}, and RotNet~\cite{gidaris2018unsupervised}.
\section{Results}\vspace{-0.25cm}
\label{sec:results}
The experimental results for linear evaluation and finetuned networks show that our method significantly outperforms the other SSL approaches. Table 1 shows the linear evaluation and finetune experiments with a shallow CNN architecture - Resnet8. In Tiny-ImageNet, the ``Patch RelNet'' with ratio 0.3 gives the best performance compared to all others in both linear and finetuned evaluations.
In STL-10, both of proposed networks significantly outperform the others. Especially, ``Patch RelNet 0.4'' is ahead the second-best method RotNet $\sim 2$\%. Table 2 compares SSL methods in a deeper backbone CNN (ResNet32).
\begin{table}[ht!]
    \centering
    \begin{tabular}{|c|c|c|} \hline
    Method & Linear & Tuned \\ \hline \hline
    Supervised & \multicolumn{2}{|c|} {74.54} \\ \hline
    Random weight & 27.19 & 60.20  \\ \hline
    Deep InfoMax ~\cite{hjelm2018learning} & 45.41 & 68.56 \\ \hline
    Deep Cluster ~\cite{caron2018deep} & 41.35 & 65.45 \\ \hline
    RotNet~\cite{gidaris2018unsupervised} & 52.25 & 76.06\\ \hline\hline
    Patch RelNet 0.3 & 50.15  & 75.55\\
    Patch RelNet 0.4 & 45.76 &  72.61\\ 
    Patch RelNet 0.5 & 42.38 & 73.70 \\ 
    Patch RotNet 0.3 & 48.75 & 68.71 \\ 
    \textbf{Patch RotNet 0.4} & \textbf{52.63} & \textbf{77.06} \\ 
    \textbf{Patch RotNet 0.5} & 50.05 & \textbf{78.38}\\ \hline
    \end{tabular}
    \caption{Test accuracy. Self-supervised training on unlabeled data with linear evaluation and finetuning on labeled data. Results of different methods on dataset STL-10 with a deeper backbone (\textcolor{blue}{ResNet-32}). \textbf{Best results are highlighted in bold}.}
    \label{tab:resnet32}\vspace{-0.1cm}
\end{table}
Our method achieves state-of-the-art performance with 2.32\% improvement compared to finetuned RotNet. Our approach also demonstrates its effectiveness in learning general-purpose features by outperforming the fully-supervised learning method. Furthermore, Fig.~\ref{fig:res_3} shows that the proposed network understands what changes happen in the local areas and puts more attention there. Additionally, Fig.~\ref{fig:res_4} illustrates the t-SNE scatter of our method and RotNet. The features were extracted from the finetuned models of RotNet and ``Patch RotNet'' gives accuracies of 76.06\% and 78.38\%. The visualized results show capability of the proposed ``Patch RotNet'' in clustering separably the images.\vspace{-0.3cm}

\section{Conclusion}
\label{sec:conclusion}
This paper proposes a novel self-supervised learning scheme that explores the local features to recognize the visual changes in a local part of a given image. Our approach encourages the networks to understand both the local and global-context within images in order to distinguish them. The proposed method facilitates learning comprehensive general-purpose visual representations, which could be used in the downstream tasks. The backbone networks which were trained using our methodology outperform the previous works on both of two challenging Tiny-ImageNet and STL-10 datasets. 

\vfill \pagebreak \newpage
\label{sec:ref}
\bibliographystyle{IEEEbib}
\bibliography{strings,refs}

\begin{thebibliography}{10}

\bibitem{deng2009imagenet}
Jia Deng, Wei Dong, Richard Socher, Li-Jia Li, Kai Li, and Li~Fei-Fei,
\newblock ``Imagenet: A large-scale hierarchical image database,''
\newblock in {\em Proceedings of the Conference on Computer Vision and Pattern
  Recognition}. IEEE, 2009, pp. 248--255.

\bibitem{bachman2019learning}
Philip Bachman, Devon Hjelm, and William B,
\newblock ``Learning representations by maximizing mutual information across
  views,''
\newblock in {\em Advances in Neural Information Processing Systems}, 2019, pp.
  15535--15545.

\bibitem{gidaris2018unsupervised}
Spyros Gidaris, Praveer Singh, and Nikos Komodakis,
\newblock ``Unsupervised representation learning by predicting image
  rotations,''
\newblock in {\em International Conference on Learning Representations}, 2018.

\bibitem{doersch2015unsupervised}
Carl Doersch, Abhinav Gupta, and Alexei~A Efros,
\newblock ``Unsupervised visual representation learning by context
  prediction,''
\newblock in {\em Proceedings of the Conference on Computer Vision and Pattern
  Recognition}, 2015, pp. 1422--1430.

\bibitem{zhang2016colorful}
Richard Zhang, Phillip Isola, and Alexei~A Efros,
\newblock ``Colorful image colorization,''
\newblock in {\em Proceedings of the European Conference on Computer Vision}.
  Springer, 2016, pp. 649--666.

\bibitem{caron2018deep}
Mathilde Caron, Piotr Bojanowski, Armand Joulin, and Matthijs Douze,
\newblock ``Deep clustering for unsupervised learning of visual features,''
\newblock in {\em Proceedings of the European Conference on Computer Vision},
  2018, pp. 132--149.

\bibitem{hjelm2018learning}
Devon Hjelm, Alex Fedorov, Karan Grewal, Phil Bachman, Adam Trischler, and
  Yoshua Bengio,
\newblock ``Learning deep representations by mutual information estimation and
  maximization,''
\newblock in {\em International Conference on Learning Representations}, 2019.

\bibitem{wang2020self}
Jiangliu Wang, Jianbo Jiao, and Yun-Hui Liu,
\newblock ``Self-supervised video representation learning by pace prediction,''
\newblock in {\em Proceedings of the European Conference on Computer Vision}.
  Springer, 2020, pp. 504--521.

\bibitem{Tschannen_2020_CVPR}
Michael Tschannen, Josip Djolonga, Marvin Ritter, Aravindh Mahendran, Neil
  Houlsby, Sylvain Gelly, and Mario Lucic,
\newblock ``Self-supervised learning of video-induced visual invariances,''
\newblock in {\em Proceedings of the Conference on Computer Vision and Pattern
  Recognition}, 2020.

\bibitem{vondrick2018tracking}
Carl Vondrick, Abhinav Shrivastava, Alireza Fathi, and Kevin Murphy,
\newblock ``Tracking emerges by colorizing videos,''
\newblock in {\em Proceedings of the European Conference on Computer Vision},
  2018, pp. 391--408.

\bibitem{alwassel2020self}
Humam Alwassel, Dhruv Mahajan, Bruno Korbar, Lorenzo Torresani, Bernard Ghanem,
  and Du~Tran,
\newblock ``Self-supervised learning by cross-modal audio-video clustering,''
\newblock {\em Advances in Neural Information Processing Systems}, vol. 33,
  2020.

\bibitem{zhao2018sound}
Hang Zhao, Chuang Gan, Andrew Rouditchenko, Carl Vondrick, Josh McDermott, and
  Antonio Torralba,
\newblock ``The sound of pixels,''
\newblock in {\em Proceedings of the European Conference on Computer Vision},
  2018, pp. 570--586.

\bibitem{gao2018learning}
Ruohan Gao, Rogerio Feris, and Kristen Grauman,
\newblock ``Learning to separate object sounds by watching unlabeled video,''
\newblock in {\em Proceedings of the European Conference on Computer Vision},
  2018, pp. 35--53.

\bibitem{zhang2017split}
Richard Zhang, Phillip Isola, and Alexei~A Efros,
\newblock ``Split-brain autoencoders: Unsupervised learning by cross-channel
  prediction,''
\newblock in {\em Proceedings of the Conference on Computer Vision and Pattern
  Recognition}, 2017, pp. 1058--1067.

\bibitem{larsson2017colorization}
Gustav Larsson, Michael Maire, and Gregory Shakhnarovich,
\newblock ``Colorization as a proxy task for visual understanding,''
\newblock in {\em Proceedings of the Conference on Computer Vision and Pattern
  Recognition}, 2017, pp. 6874--6883.

\bibitem{mundhenk2018improvements}
Nathan Mundhenk, Daniel Ho, and Barry~Y Chen,
\newblock ``Improvements to context based self-supervised learning,''
\newblock in {\em Proceedings of the European Conference on Computer Vision},
  2018, pp. 9339--9348.

\bibitem{everingham2010pascal}
Mark Everingham, Luc Van~Gool, Christopher~KI Williams, John Winn, and Andrew
  Zisserman,
\newblock ``The pascal visual object classes (voc) challenge,''
\newblock {\em International journal of computer vision}, vol. 88, no. 2, pp.
  303--338, 2010.

\bibitem{noroozi2016unsupervised}
Mehdi Noroozi and Paolo Favaro,
\newblock ``Unsupervised learning of visual representations by solving jigsaw
  puzzles,''
\newblock in {\em Proceedings of the European Conference on Computer Vision}.
  Springer, 2016, pp. 69--84.

\bibitem{krizhevsky2009learning}
Alex Krizhevsky,
\newblock ``Learning multiple layers of features from tiny images,''
\newblock {\em .}, 2009.

\bibitem{zhou2014learning}
Bolei Zhou, Agata Lapedriza, Jianxiong Xiao, Antonio Torralba, and Aude Oliva,
\newblock ``Learning deep features for scene recognition using places
  database,''
\newblock {\em Advances in Neural Information Processing Systems}, vol. 27, pp.
  487--495, 2014.

\bibitem{coates2011analysis}
Adam C, Andrew Ng, and H~Lee,
\newblock ``An analysis of single-layer networks in unsupervised
  feature-learning,''
\newblock in {\em Proceedings of the 14th international conference on
  artificial intelligence and statistics}, 2011, pp. 215--223.

\end{thebibliography}
\end{document}